\documentclass{article}
\usepackage{spconf}
\usepackage{amsmath,graphicx}

\usepackage{fancyhdr}

\usepackage{comment}
\usepackage{acronym}
\usepackage{caption}
\usepackage{subcaption}
\usepackage{booktabs}
\usepackage{xcolor}
\usepackage{amssymb}
\usepackage{hyperref}

\graphicspath{ {./images/} }

\acrodef{hog}[HOG]{Histogram of Oriented Gradients}
\acrodef{gan}[GAN]{Generative Adversarial Network}
\acrodef{lbp}[LBP]{Local Binary Patterns}
\acrodef{glcm}[GLCM]{Gray-Level Co-occurrence Matrix}
\acrodef{svm}[SVM]{Support Vector Machine}
\acrodef{vit}[ViT]{Vision Transformer}
\acrodef{cnn}[CNN]{Convolutional Neural Network}
\acrodef{ssim}[SSIM]{Structural Similarity Index}
\acrodef{rf}[RF]{Random Forest}
\acrodef{iou}[IoU]{Intersection over Union}
\acrodef{ioma}[IOMA]{Intersection over Minimum Area}
\acrodef{map}[mAP]{Mean Average Precision}

\title{Defect Detection in Tire X-Ray Images: Conventional Methods Meet Deep Structures}

\twoauthors
 {Andrei Cozma$^1$, Landon Harris$^1$, Hairong Qi}
	{Department of Electrical Engineering \\ and Computer Science \\
	University of Tennessee, Knoxville, USA}
 {Ping Ji, Wenpeng Guo, Song Yuan}
	{Sailun Group Co. Ltd. \\ Qingdao, China \\}

\begin{document}

\maketitle

\footnotetext[1]{These authors contributed equally to this work.}


\thispagestyle{fancy}
\fancyhf{} 
\renewcommand{\headrulewidth}{0pt} 
\fancyfoot[C]{
\small 
\textit{
\copyright\ 2024 IEEE. Personal use of this material is permitted. Permission
from IEEE must be obtained for all other uses, in any current or future
media, including reprinting/republishing this material for advertising or
promotional purposes, creating new collective works, for resale or
redistribution to servers or lists, or reuse of any copyrighted
component of this work in other works.
}}
\renewcommand{\footrulewidth}{0.4pt} 


\begin{abstract}

This paper introduces a robust approach for automated defect detection in tire X-ray images by harnessing traditional feature extraction methods such as \ac{lbp} and \ac{glcm} features, as well as Fourier and Wavelet-based features,
complemented by advanced machine learning techniques. Recognizing the challenges inherent in the complex patterns and textures of tire X-ray images, the study emphasizes the significance of feature engineering to enhance the performance of defect detection systems.
By meticulously integrating combinations of these features with a \ac{rf} classifier and comparing them against advanced models like YOLOv8, the research not only benchmarks the performance of traditional features in defect detection but also explores the synergy between classical and modern approaches. The experimental results demonstrate that these traditional features, when fine-tuned and combined with machine learning models, can significantly improve the accuracy and reliability of tire defect detection, aiming to set a new standard in automated quality assurance in tire manufacturing.

\end{abstract}

\begin{keywords}
tire defect, automated defect detection, x-ray imaging, 
\end{keywords}


\section{Introduction}
\label{sec:intro}

Ensuring the integrity of tire structures is crucial for vehicle safety. 
X-ray imaging provides an effective means to inspect the internal composition of tires, revealing defects that are not easily seen externally. 
However, due to the inherent challenges brought by tire X-ray images, defect inspection has been heavily relied on by human inspectors.   
The motivation for this research is to develop an automated system capable of accurately detecting and identifying various types of tire defects, thereby reducing the reliance on manual inspection. 


The drive towards automating defect detection comes from the inherent limitations of manual inspections. 
Inspectors constantly face challenges such as varying expertise and fatigue, which can result in missed defects or false identifications. 
Moreover, manual inspection is time-consuming and can slow down production lines. 
Our research is motivated by the need for a more consistent, time-efficient, and accurate approach to scrutinizing tire integrity using X-ray imagery.
\begin{figure}[t]
    \centering
    \begin{subfigure}[t]{0.3\linewidth}
        \includegraphics[width=\linewidth]{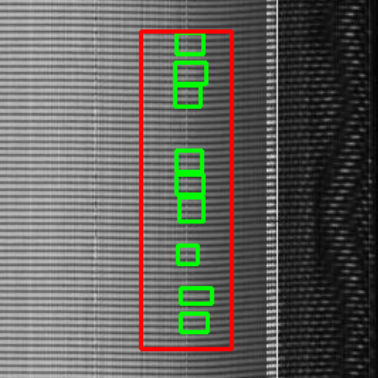}
        \caption{Blister 1}
        \label{fig:ex_blister1}
    \end{subfigure}
    \begin{subfigure}[t]{0.3\linewidth}
        \includegraphics[width=\linewidth]{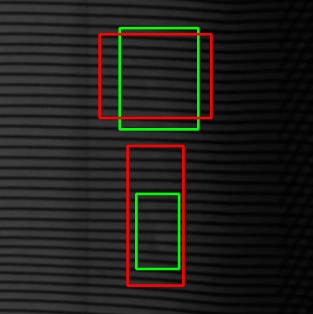}
        \caption{Blister 2}
        \label{fig:ex_blister2}
   \end{subfigure}
    %
   \begin{subfigure}[t]{0.3\linewidth}
        \includegraphics[width=\linewidth]{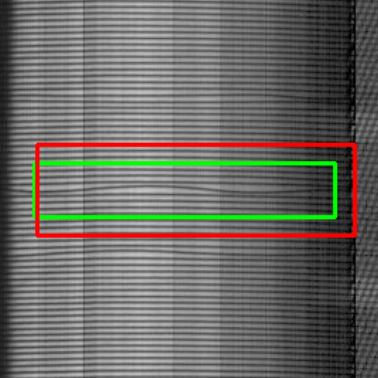}
        \caption{Wire Defect 1}
        \label{fig:ex_wire1}
    \end{subfigure}

    \caption{Examples of two major tire defect types (blister and wire) with inherent challenges. Note: annotations (green) and predictions (red) are marked.}
    \label{fig:predictions}
\end{figure}
%
Current inspection methods face significant hurdles that an automated system must overcome:
      
\textit{The High-Resolution X-Ray Samples Challenge}: 
High-resolution image processing is computationally demanding, even more so with deep learning techniques. 
Downsizing images for faster processing can lead to loss of critical details, affecting accurate defect detection. 
Resizing images to a consistent width maintains spatial consistency but can omit important information from the original, higher-resolution images.

\textit{The Defect Characteristics Challenge}: 
Tire defects come in various sizes and shapes, making them difficult to detect. For example, 
they can be as tiny as $40 \times 40$ pixels in thousands of pixels wide and tall images, as shown in Fig.~\ref{fig:ex_blister1}. 
Additionally, the defects are often difficult to discern due to their low contrast against the tire background, as shown in Fig.~\ref{fig:ex_blister2}. 
Adaptable methodologies are needed to accurately detect and identify the range of characteristics within similar defects.

\textit{The Tread Pattern Variability Challenge}: 
Detecting defects in tires is further complicated by the variability in tread patterns, as tread designs often have features that look like defects, as shown in Fig.~\ref{fig:ex_wire1}. 
Additionally, the patterns are often anisotropic, posing significant challenges for \ac{cnn} architectures that excel in analyzing isotropic features. 
This anisotropy makes it difficult for \acp{cnn} to learn and discern the necessary spatial patterns effectively. 
Therefore, it is crucial to develop specific metrics to quantify the differences within defect-free images, known as intra-class variations, and differentiate them from inter-class differences that define defective and non-defective images.

\textit{The Data Imbalance Challenge}: 
The data imbalance issue, where datasets contain more defect-free samples than defective samples, poses a risk of bias in machine learning models. 
Such models might tend towards predicting the majority class, necessitating strategies to correct this imbalance to ensure accurate defect detection.


We hypothesize that traditional engineered features can be used not only to improve the accuracy of a \ac{cnn}-based detection model but also exclusively for precise defect detection in tire X-ray images. 
By fine-tuning the balance between speed and accuracy, we aim to establish a robust solution that meets industrial standards for real-time performance without compromising quality assurance. The contribution of this paper is three-fold: 

\begin{itemize}
    \item Develop a comprehensive approach for automated detection of defects in tire X-ray images that meets both real-time and accuracy requirements. 
    \item Revisit the merit of some traditional feature extraction and classification approaches to the challenge of object detection on images with anisotropic patterns. 
    \item Explore the synergy between classical feature engineering and advanced deep learning methods to enhance detection performance.
\end{itemize}


\section{Related Works}
\label{sec:lit}

There have been a couple of recent works focusing on using deep networks to solve the tire defect detection problem. 
MSANet represents a significant step in applying neural networks specifically for radiographic tire image analysis \cite{zhao2022MSANet}. 
This method employs the YOLOv4 suite of detection algorithms \cite{bochkovskiy2020yolov4} and adds a multi-scale self-attention module to handle the anisotropic texture of the background. Adding the self-attention module addresses two issues: the anisotropic nature of the X-ray images and the need for global context to perform detection. This method, however, uses a proprietary dataset and does not make their code public, making validation of their results and comparison as a baseline challenging. 

One pivotal advancement is the application of transfer learning and domain adaptation. 
This approach implements a dual-domain adaptation-based transfer learning strategy. 
To improve performance across various X-ray scanners, \cite{zhang2022tire} utilized a ResNet and \ac{vit} architecture for feature extraction and perform domain adaptation among different X-ray scanners. 

\acp{gan} have been used in the unsupervised domain to combat the defective~/~non-defective class imbalance in real-world data. 
\cite{wang2021unsupervised} utilized \acp{gan} in an unsupervised learning context to generate a model with the task of inpainting the original image. 
\ac{ssim} and L2 distance discriminate between the original and reconstructed image were used to determine whether a defect is present. 

On the more traditional side of image processing, \acf{hog} and \acf{lbp} have been used for defect detection with the omission of deep learning altogether, instead using \acp{svm} for the classification task using standard images of the tire exterior \cite{liu2023tire}. Similarly, \cite{cui2016defect} shows the use of principal components for removing the background texture in tire X-ray images.


\section{Methodology}
\label{sec:method}


Defect detection in images is a multi-stage process that requires careful 
preparation of the image data, feature extraction, model selection, and 
training, optimization, and performance evaluation to ensure the creation 
of an accurate and robust defect detection system. 
This section first discusses the engineered features adopted in the automated detection system. 
We then elaborate on leveraging the advantages of both engineered features and deep network structures for further performance improvement. 


\subsection{Feature Extraction}
\label{subsec:feature}


We have strategically chosen a set of traditional feature extraction techniques to identify defects within tire X-ray images. Each feature is selected for its ability to capture specific image attributes crucial for our analysis. The following is a breakdown of our rationale behind choosing these features and our expectations regarding their performance in this context.

\textit{\acf{lbp}}. 
\acp{lbp} are traditionally utilized for texture classification \cite{lbp1}.
They excel in identifying local texture patterns by comparing a pixel with its surrounding neighbors and encoding this relationship as a binary number \cite{lbp1}.
We employ uniform LBP with radii of 
2, 8, and 16, and the number of points calculated as $\text{points} = \text{radius} \times 8$,
anticipating these configurations to comprehensively capture textural variations at multiple scales.
We hypothesize that LBPs could detect textural anomalies in the tire X-ray images.
Blisters and wire defects disrupt the regular textural pattern of a tire, and LBPs, especially Uniform and Rotation-Invariant variants, are expected to be sensitive to these disruptions.
For the traditional method, the LBP histograms, with the number of bins calculated as $\text{bins} = \text{points} \times (\text{points} - 1) + 3$,
provide a statistical view of these textural irregularities.
Besides the LBP histogram, we additionally extract various statistical features from these LBPs, including mean, median, min, max, standard deviation, and energy.
These combined LBP features are anticipated to be crucial in distinguishing between defective and non-defective areas.
    
\textit{\acf{glcm}}. 
\ac{glcm} has been a go-to method for texture analysis in fields like medical imaging. 
It assesses the spatial relationships between pixels by analyzing how often pairs of pixels with specific values appear in one particular spatial orientation \cite{glcm1}.
We hypothesize that in our application, GLCM could offer valuable insights into tire surface uniformity and textural variation. 
We expect that GLCM-derived features like contrast, dissimilarity, homogeneity, energy, and correlation will be instrumental in differentiating between the regular texture of a healthy tire and the irregular patterns indicative of blisters and wire defects. 
The chosen distances and angles for GLCM analysis are intended to capture these textural properties over various scales and orientations.


\textit{Wavelet Features}
Wavelet transforms are famed for decomposing signals into frequency components across multiple scales. 
Their use spans signal processing and image compression due to their ability to localize both time (or space) and frequency components.
We expect wavelet features to excel in identifying defects manifesting at different scales and orientations for tire X-ray images. 
The Haar wavelet’s capability to decompose images into approximation and detail coefficients could be critical in pinpointing subtle variances caused by defects like blisters or wire deformities. 
We extract various statistical features from each level of detail of the wavelet decomposition, including mean, median, min, max, standard deviation, and energy.
    

\textit{Fourier Features}. The Fourier Transform is instrumental in translating spatial data into the frequency domain, revealing underlying frequency components. 
In the context of tire X-ray images, we hypothesize that the Fourier Transform can unveil defects through altered frequency patterns. 
Defects like blisters and wire deformities are expected to introduce unique frequency signatures different from the normative tire texture. 
Specifically, besides extracting the set of statistical measures, we also extract a specialized set of spectral features, including spectral centroid, 
spectral bandwidth, 
spectral flatness, 
and spectral roll-off. 


Our selection of 
features 
is rooted in the belief that each will provide a unique lens through which to examine and interpret the tire X-ray images. 
From capturing local textural disruptions to understanding global patterns and variances, these features collectively form a comprehensive toolkit for identifying and classifying tire defects with precision and reliability.


\subsection{YOLO with Augmented Features}

In addition to the investigation of classical feature engineering approaches, we further study the potential benefit of integrating these advanced features into the deep learning framework. 

\begin{figure}[th]
    \centering
    \begin{subfigure}[t]{0.32\linewidth}
        \centering
        \includegraphics[width=\linewidth]{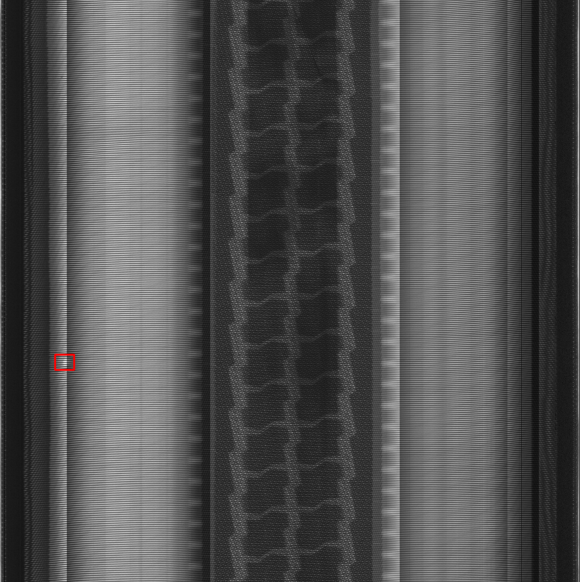}    
        \caption{Original Image}
        \label{sfig:greyscale}
    \end{subfigure}
    \begin{subfigure}[t]{0.32\linewidth}
        \centering
        \includegraphics[width=\linewidth]{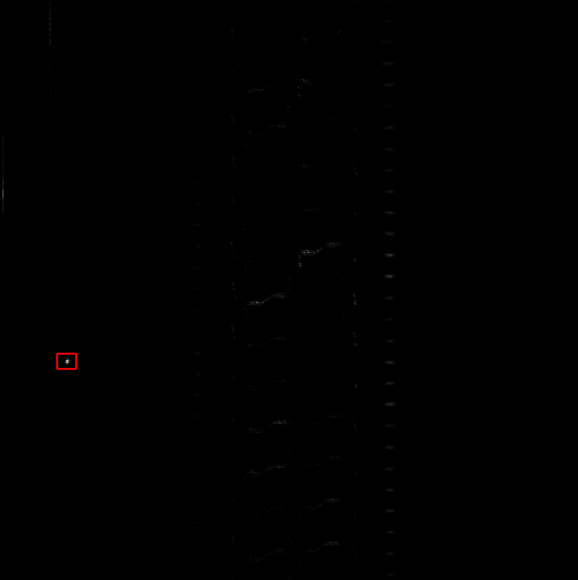}
        \caption{Background Removed}
        \label{sfig:background_removed}
    \end{subfigure}
    \begin{subfigure}[t]{0.32\linewidth}
        \centering
        \includegraphics[width=\linewidth]{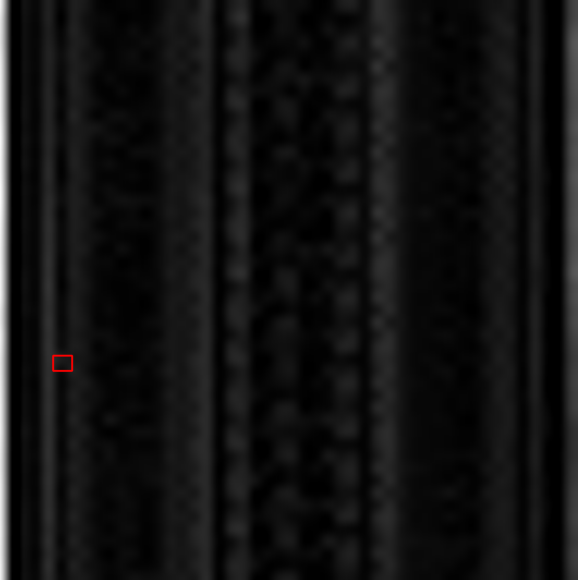}
        \caption{Wavelet Reconstruction}
        \label{sfig:wavelet_reconstruction}
    \end{subfigure}
    \caption{The three channels used as input for the YOLO detection model.}

    \label{fig:3channel_yolo}
\end{figure}

We use YOLOv8 as the baseline deep model, a member of the YOLO family of detection models for its real-time performance and its incorporation of both channel attention and spatial attention \cite{yolov8}. 


Since YOLO is designed to work with 3-channel images (although it is flexible to handle any number of channels), we choose to integrate the engineered features discussed in Sec.~\ref{subsec:feature} by augmenting these features to the original grayscale image as an additional input channel. Fig.~\ref{fig:3channel_yolo} shows such an example of augmenting the original image (Fig.~\ref{sfig:greyscale}) with a texture removal process (Fig.~\ref{sfig:background_removed}) and Wavelet reconstruction (Fig.~\ref{sfig:wavelet_reconstruction}). The texture removal process iterates through horizontal slices of the image, identifying the top-K most similar slices for each slice. The average of these top-K slices is then subtracted from the original slice to remove the background. The background-removed image is exponentiated to enhance the separation between the background and foreground. This process is repeated for multiple-sized slices, resulting in an image with the defect highlighted.

\subsection{Model Selection, Optimization, and Pre-Processing}
\label{subsec:model}

For the traditional method, selecting the \acf{rf} Classifier is primarily motivated by its established effectiveness and balance between accuracy and computational efficiency \cite{breiman2001random}. A grid search with 5-fold cross-validation was utilized to systematically optimize the hyperparameters for the model, including the number of trees, criterion functions, and maximum features to consider at each split. 




The image pre-processing phase for training the \ac{rf} begins with normalizing the pixel values to a range of 0 to 255 to ensure uniform luminance levels. This is necessary to maintain consistency in the feature extraction phase. 
The image dimensions are also adjusted based on predetermined window and step sizes, ensuring that a sliding window can cover the entire image seamlessly without leaving any areas untouched. 
This involves modifying the image width to align with the step size and readjusting the height to maintain the aspect ratio.
Finally, a slight Gaussian blur is applied to reduce noise and improve image quality, eliminating noise while preserving essential features. 


For the YOLO model, since the images are of variable resolution and contain tiny defects relative to the total size of the image, a windowed training and inference approach is used. 
In this phase, a systematic scan of the entire image is conducted using square windows of a predetermined size, advancing each time by a specified step size, both of which are tunable hyper-parameters. The image is split into windows of size $200 \times 200$, and YOLOv8-medium is trained on a mix of windows and the entire image. During inference, \cite{akyon2022slicing} is used to slice the image into overlapping patches, run inference through the YOLO model, and merge the resulting detections. Images are resized to a consistent width. For training, the images are cropped into $448 \times 448$px windows with a stride of 128px, including only 15\% of the crops without defects. During inference, the images are sliced at runtime into crops of similar size, and a stride is dynamically calculated to cover the entire image’s width and height. 


\subsection{The Probability Map Ensemble}
\label{subsec:eval}

The object detection process begins with the images undergoing pre-processing 
as described in the previous section.
These images are then dissected into smaller square windows, where the size and stride configuration are determined by the \textit{Window Size} and \textit{Step} hyperparameters that were previously selected for the corresponding model during training.

Simultaneously, a probability mask is constructed, resembling a 3-dimensional overlay on the original image. 
This mask’s dimensions mirror the image’s height and width, with an added depth representing the number of classes the model discerns.



Each window extracted from the pre-processed images is then scrutinized by the \ac{rf} classifier to deduce the most probable class it belongs to, alongside a confidence score reflecting the degree of certainty behind the classification. These confidence scores are methodically accumulated in the corresponding class layer of the probability mask, ensuring every window’s prediction contributes to the overall assessment of each pixel’s class affiliation.

As windows frequently overlap, a single pixel’s final classification is influenced by multiple predictions. 
To harmonize these overlapping predictions, the amassed probability scores undergo a normalization step where each score is adjusted by the power of 2.8, intensifying the distinction between high and low confidence areas in detecting defects. 
Following this enhancement, the background mask’s probabilities are subtracted from the defect class masks. 
This critical manipulation reduces the probabilities in regions confidently identified as background, thereby nullifying their impact on the detection of defects. 
To isolate the regions with the highest likelihood of containing defects and finalize the heatmap for each defect class, a threshold is applied, discarding probabilities below the 0.98 quantile of the distribution obtained from the defect masks — a decisive move that effectively filters out the most significant and probable defect areas from the cumulative insights gained through window-wise analysis.






\section{Experiments and Results}
\label{sec:exp_res}

We conduct three sets of experiments to evaluate the proposed strategies thoroughly. The first set compares traditional feature-based detection and YOLO with augmented features on the detection accuracy. The second set evaluates the effectiveness of different features. The third set analyzes the effect of hyperparameters.

\subsection{Dataset Overview}
\label{subsec:dataset}

Our dataset is collected from Sailun Group Co. Ltd. with the intention of making it publically accessible. It consists of 1054 tire X-ray scans presented as grayscale images, annotated with bounding boxes to highlight two types of defects: blisters and wire defects. 

The X-ray scans vary widely in size, spanning widths from 1536 to 3328 pixels (median of 2469px) and heights from 1625 to 14600 pixels (median of 7777px).
The bounding boxes delineating the defects are relatively small compared to the scans’ dimensions, with a median size of 48x39px, corresponding to 1.9\% and 0.5\% of the median scan width and height, respectively. 
The dimensions of wire defect annotations generally remain compact. In contrast, blister annotations display a high degree of variability—some as tall as 2000 pixels—indicating the presence of long and slender blisters. 
This marked diversity in the size and appearance of the defects poses significant challenges to the development of practical detection algorithms.



\subsection{Metrics}
Because our method is designed to be an indication to an operator of areas likely to contain defects, we do not evaluate with \ac{iou} as in \cite{zhao2022MSANet,bochkovskiy2020yolov4,yolov8}. Instead, we consider a true positive to be a prediction that contains at least one detection. To balance true positives and false negatives, we multiply each true positive by the number of detections it covers. A false negative is any defect not covered by at least 40\% of a prediction. Finally, a false positive is a prediction that covers no defects. 

For the \ac{rf} classifier method, each square window must be assigned a ground truth label based on the annotations present in our dataset. 
To quantify the amount of overlap between the sliding window and the ground truth bounding boxes we use the \ac{ioma} metric. 
The label of the ground truth annotation with the highest \ac{ioma} value is assigned as the label for the window. 
If a window does not overlap with any annotations, it is labeled as “background,” indicating no defect in that window. 
For the \textit{Window Size} hyperparameter, we chose a default value of 128px; for the \textit{Step Size}, a default of 32px; and for the \textit{\ac{ioma} Threshold}, a default of 0.1. We further analyze the effects of these hyper-parameters on various per-window classification metrics.





\subsection{Experiment 1: Comparing Engineered Features vs. Deep Models}

In this first set of experiments, we conduct a comprehensive evaluation of the proposed engineered feature-based detection approaches and their deep model counterparts. The results are shown in Table \ref{tab:main_results}. We observe a clear performance gain using engineered features as compared to YOLO-based models. This is largely due to the imbalanced training set, the large variation in defect size and shape, and the sometimes extremely high intra-class difference but low inter-class difference. 

To evaluate the YOLO models, we use the standard object-detection evaluation metrics where an object is a true positive when the label and detection have an IoU greater than some threshold. While the threshold in standard evaluation methods is typically 0.50 to 0.75, we choose to evaluate at 0.20 to promote a fair comparison to the random forest method, which will inherently have a lower IoU. 

\begin{table*}
    \centering
    \caption{Comparison between traditional engineered feature frameworks (bottom section) and YOLO-based frameworks in precision and recall.
    \textbf{BR}:~Background Removal, \textbf{WR}:~Wavelet Reconstruction, \textbf{L}:~\ac{lbp} Features, \textbf{G}:~\ac{glcm} Features, \textbf{F}:~Fourier Features, \textbf{W}:~Wavelet Features. }
    \label{tab:main_results}
    \begin{tabular}{@{\extracolsep{6pt}}lcccccc@{}}  
        \toprule
        \multicolumn{1}{c}{} & \multicolumn{2}{c}{Blister} & \multicolumn{2}{c}{Wire} \\\cline{2-3}\cline{4-5} 
        Method                   & Prec(\%) & Rec(\%) & Prec(\%) & Rec(\%) & F1 & IoU \\ 
        \midrule
        YOLOv8 Baseline          & 0.190        & 0.373          & 0.151          & 0.506          & 0.240          & 0.20 \\
        YOLOv8 Augmented (BR,WR) & 0.291        & \textbf{0.902} & 0.274          & \textbf{0.885} & 0.422          & 0.20 \\
        YOLOv8 Augmented (BR)    & 0.262        & 0.875          & 0.247          & 0.856          & 0.393          & 0.20\\
        YOLOv8 Augmented (WR)    & 0.238        & 0.853          & 0.225          & 0.837          & 0.353          & 0.20 \\ 
        \midrule
        Random Forest (L,W)     & 0.633         & 0.714          & \textbf{0.564} & 0.636          & 0.634          & \textbf{0.161} \\ 
        Random Forest (G,F,W)   & 0.741         & 0.766          & 0.562          & 0.651          & 0.678          & 0.128 \\
        Random Forest (G,W)     &\textbf{0.759} & 0.766          & 0.554          & 0.662          & \textbf{0.683} & 0.083 \\
        \bottomrule
    \end{tabular}
\end{table*}




\subsection{Experiment 2: Ablation Study on the Feature Sets}
\label{subsec:eff_feature}

In this set of experiments, we examine our defect detection framework’s components in-depth, isolating and evaluating each to understand their individual contributions to the model’s overall performance. 
Specifically, we dissect the roles played by various feature sets, including \ac{lbp}, \ac{glcm}, Fourier, and Wavelet features. 

Unlike the previous sections, where we utilized combined metrics to gauge the effectiveness of our entire pipeline, here we focus on per-window classification metrics, measuring precision, recall, and F1 scores directly from the predictions of our \ac{rf} classifier. 
This approach allows us to capture a granular view of how well the classifier detects the defects at the window level before any post-processing into a comprehensive probability heatmap or mask. 


We make the following interesting observations. 
First, excluding \ac{lbp} and combining \ac{glcm}, Fourier, and Wavelet features resulted in the most optimal performance, with top scores in precision (0.922), recall (0.918), and F1-score (0.918). 
This combination evidences a robust synergistic effect beneficial for achieving high accuracy in per-window image classification.
    
Second, utilizing \ac{lbp} as the sole feature yielded the lowest performance metrics, underscoring its inadequacy for standalone use in per-window classification. 
Moreover, including \ac{lbp} in feature set combinations typically diminished performance, suggesting that it might introduce redundancy or reduce the contribution of other features in such contexts.
    
Finally, including Wavelet features—especially when excluding \ac{lbp}—consistently improved classification performance. 
This highlights the importance of Wavelet features in effectively capturing critical details necessary for accurate per-window classification.

\begin{table}[t]
    \centering
    \caption{ Results showcasing the effects of various feature set combinations }
    \label{tab:eff_feature}
    \begin{tabular}{lllllll}
    \toprule
           L &        G &        F &        W & Precision & Recall & F1 \\
    \midrule
    $\times$ & $\times$ & $\times$ & $\times$ &               0.911 &            0.889 &              0.897 \\
    $\times$ & $\times$ & $\times$ &          &               0.905 &            0.873 &              0.886 \\
    $\times$ & $\times$ &          & $\times$ &               0.905 &            0.879 &              0.889 \\
    $\times$ & $\times$ &          &          &               0.897 &            0.858 &              0.872 \\
    $\times$ &          & $\times$ & $\times$ &               0.892 &            0.853 &              0.867 \\
    $\times$ &          & $\times$ &          &               0.871 &            0.811 &              0.831 \\
    $\times$ &          &          & $\times$ &               0.889 &            0.846 &              0.861 \\
    $\times$ &          &          &          &               0.862 &            0.795 &              0.816 \\
             & $\times$ & $\times$ & $\times$ &      \textbf{0.922} &   \textbf{0.918} &     \textbf{0.918} \\
             & $\times$ & $\times$ &          &               0.914 &            0.907 &              0.909 \\
             & $\times$ &          & $\times$ &               0.919 &            0.914 &              0.916 \\
             & $\times$ &          &          &               0.903 &            0.891 &              0.895 \\
             &          & $\times$ & $\times$ &               0.901 &            0.893 &              0.895 \\
             &          & $\times$ &          &               0.794 &            0.752 &              0.766 \\
             &          &          & $\times$ &               0.898 &            0.892 &              0.894 \\
    \bottomrule
    \end{tabular}
\end{table}


\subsection{Experiment 3: Effect of Dataset Generation Parameters}
\label{subsec:eff_data}

In this set of experiments, we aim to assess the per-window classification performance of \ac{rf} models, emphasizing how variations in the dataset construction parameters—window sizes, step sizes, and thresholds—affect the models’ accuracy in defect detection tasks. 
Our analysis focuses on comparing models trained with window sizes of 128, 256, and 384 pixels, step sizes of 32 and 64 pixels, and \ac{ioma} thresholds of 0.1 and 0.3 for ground truth overlap to identify the optimal configurations for maximizing the macro average precision, recall, and F1 scores across classes. 
Moreover, we also integrate the previously determined optimal feature set combination: \ac{glcm}, Fourier, and Wavelet features.

The results highlight a direct correlation between increased window size and improved model performance, with a notable peak in precision (0.973), recall (0.980), and F1 score (0.976) for the model utilizing a 384-pixel window, a 32-pixel step size, and a 0.3 threshold. 
This configuration demonstrates the advantage of larger windows in providing more contextual information, which is crucial for accurate defect detection. 
Conversely, the analysis also reveals that while larger windows enhance accuracy, optimizing step size and threshold parameters is equally imperative. 
A step size of 32 pixels, compared to 64, ensures denser coverage and a higher likelihood of defect capture. 
In contrast, a threshold of 0.3 strikes a suitable balance by including windows with substantial defect overlap, effectively reducing false positives without overlooking minor defects.

\begin{table}
    \centering
    \caption{ Results showcasing the effects of Window Size, Step, and Threshold }
    \label{tab:eff_data}
    \begin{tabular}{rrrlll}
    \toprule
     Window &  Step &  Thresh & Precision & Recall & F1 \\
    \midrule
           128 &      32 &          0.1 &               0.919 &            0.916 &              0.916 \\
           128 &      32 &          0.3 &               0.931 &            0.926 &              0.927 \\
           128 &      64 &          0.1 &               0.841 &            0.808 &              0.819 \\
           128 &      64 &          0.3 &               0.851 &            0.804 &              0.821 \\
           256 &      32 &          0.1 &               0.958 &            0.965 &              0.961 \\
           256 &      32 &          0.3 &               0.962 &            0.968 &              0.964 \\
           256 &      64 &          0.1 &               0.916 &            0.915 &              0.913 \\
           256 &      64 &          0.3 &               0.926 &            0.926 &              0.924 \\
           384 &      32 &          0.1 &               0.970 &            0.977 &              0.973 \\
           384 &      32 &          0.3 &      \textbf{0.972} &   \textbf{0.979} &     \textbf{0.976} \\
           384 &      64 &          0.1 &               0.936 &            0.945 &              0.939 \\
           384 &      64 &          0.3 &               0.935 &            0.948 &              0.939 \\
    \bottomrule
    \end{tabular}
\end{table}



\section{Discussion}
\label{sec:discuss}

This study has demonstrated the viability of combining traditional feature extraction techniques with advanced machine learning models to enhance the performance of automated defect detection systems in tire X-ray images. The experiments conducted reveal that a mix of \ac{glcm}, Fourier, and Wavelet features, when used in conjunction with an \ac{rf} classifier, significantly outperforms other combinations, including those that utilize \ac{lbp}. This finding underscores the value of carefully selected feature sets in improving the accuracy of defect detection systems.

Our exploration into optimizing dataset generation parameters, such as window size, step size, and the threshold for determining ground truth overlap, has provided deeper insights into effective data preparation. These insights emphasize the importance of balancing sufficient contextual information against the risk of increasing false positives. Furthermore, the comparison with state-of-the-art models like YOLOv8 highlights the competitive edge that can be achieved through a strategic blend of traditional feature engineering and machine learning. Even in an era heavily dominated by deep learning solutions, our approach illustrates that traditional feature extraction methods hold significant value. They should not be overlooked but rather integrated with contemporary technology to tackle complex image analysis effectively.


\section{Conclusions}
\label{sec:conclusion}

In conclusion, our research presents a compelling argument for integrating traditional feature extraction with machine learning algorithms to create robust and efficient automated defect detection systems for tire X-ray images. The refinement in feature selection and optimization of dataset parameters have paved the way for our framework to not only match but, in certain aspects, surpass the capabilities of deep learning models like YOLOv8. Future work will expand this framework’s applicability and further refine its performance, indicating a promising direction for combining classical and modern approaches in industrial quality control and beyond.

Looking forward, the results of this study advocate for a hybrid approach to machine vision, emphasizing the untapped potential of combining traditional feature engineering with advanced machine learning techniques. By continuing to explore these methods, we anticipate substantial advancements in automated defect detection systems, offering new perspectives in industrial automation and quality assurance practices.


\bibliographystyle{IEEEbib}
\bibliography{refs}

\begin{thebibliography}{10}

\bibitem{zhao2022MSANet}
Mengmeng Zhao, Zhouzhou Zheng, Yingwei Sun, Yankang Chang, Chengliang Tian, and Yan Zhang,
\newblock ``{{MSANet}}: Efficient detection of tire defects in radiographic images,''
\newblock {\em Measurement Science and Technology}, vol. 33, no. 12, pp. 125401, Sept. 2022.

\bibitem{bochkovskiy2020yolov4}
Alexey Bochkovskiy, Chien-Yao Wang, and Hong-Yuan~Mark Liao,
\newblock ``{{YOLOv4}}: {{Optimal Speed}} and {{Accuracy}} of {{Object Detection}},'' Apr. 2020.

\bibitem{zhang2022tire}
Yulong Zhang, Yilin Wang, Zhiqiang Jiang, Li~Zheng, Jinshui Chen, and Jiangang Lu,
\newblock ``Tire {{Defect Detection}} by {{Dual-Domain Adaptation-Based Transfer Learning Strategy}},''
\newblock {\em IEEE Sensors Journal}, vol. 22, no. 19, pp. 18804--18814, Oct. 2022.

\bibitem{wang2021unsupervised}
Yilin Wang, Yulong Zhang, Li~Zheng, Liedong Yin, Jinshui Chen, and Jiangang Lu,
\newblock ``Unsupervised {{Learning}} with {{Generative Adversarial Network}} for {{Automatic Tire Defect Detection}} from {{X-ray Images}},''
\newblock {\em Sensors}, vol. 21, no. 20, pp. 6773, Jan. 2021.

\bibitem{liu2023tire}
Hongbin Liu, Xinghao Jia, Chenhui Su, Hongjuan Yang, and Chengdong Li,
\newblock ``Tire appearance defect detection method via combining {{HOG}} and {{LBP}} features,''
\newblock {\em Frontiers in Physics}, vol. 10, 2023.

\bibitem{cui2016defect}
Xuehong Cui, Yun Liu, and Chuanxu Wang,
\newblock ``Defect automatic detection for tire x-ray images using inverse transformation of principal component residual,''
\newblock in {\em 2016 Third International Conference on Artificial Intelligence and Pattern Recognition (AIPR)}, Sept. 2016, pp. 1--8.

\bibitem{lbp1}
T.~Ojala, M.~Pietikainen, and T.~Maenpaa,
\newblock ``Multiresolution gray-scale and rotation invariant texture classification with local binary patterns,''
\newblock {\em IEEE Transactions on Pattern Analysis and Machine Intelligence}, vol. 24, no. 7, pp. 971--987, July 2002.

\bibitem{glcm1}
Robert~M. Haralick, K.~Shanmugam, and Its'Hak Dinstein,
\newblock ``Textural features for image classification,''
\newblock {\em IEEE Transactions on Systems, Man, and Cybernetics}, vol. SMC-3, no. 6, pp. 610--621, Nov 1973.

\bibitem{yolov8}
Glenn Jocher, Ayush Chaurasia, and Jing Qiu,
\newblock ``{Ultralytics YOLO},'' Jan. 2023.

\bibitem{breiman2001random}
Leo Breiman,
\newblock ``Random forests,''
\newblock {\em Machine Learning}, vol. 45, no. 1, pp. 5--32, Oct. 2001.

\bibitem{akyon2022slicing}
Fatih~Cagatay Akyon, Sinan Onur~Altinuc, and Alptekin Temizel,
\newblock ``Slicing aided hyper inference and fine-tuning for small object detection,''
\newblock in {\em 2022 IEEE International Conference on Image Processing (ICIP)}, Oct. 2022, pp. 966--970.

\end{thebibliography}

\end{document}